\documentclass[10pt,twocolumn,letterpaper]{article}

\usepackage{wacv}              % To produce the CAMERA-READY version
\usepackage[accsupp]{axessibility} % Improves PDF readability for those with disabiliti
\usepackage{graphicx}
\usepackage{amsmath}
\usepackage{amssymb}
\usepackage{booktabs}
\usepackage{graphicx}
\usepackage{booktabs}
\usepackage{tabularx}
\usepackage{multirow}
\usepackage{pifont}
\usepackage{amssymb}
\usepackage{color}
\usepackage{enumitem}

\usepackage[pagebackref,breaklinks,colorlinks]{hyperref}

\usepackage[capitalize]{cleveref}
\crefname{section}{Sec.}{Secs.}
\Crefname{section}{Section}{Sections}
\Crefname{table}{Table}{Tables}
\crefname{table}{Tab.}{Tabs.}

\def\wacvPaperID{1064} % *** Enter the WACV Paper ID here
\def\confName{WACV}
\def\confYear{2025}

\begin{document}

\title{\emph{UCDR-Adapter:} Exploring Adaptation of Pre-Trained Vision-Language Models for Universal Cross-Domain Retrieval}

\author{
    Haoyu Jiang\textsuperscript{1}\thanks{Part of this work was completed at Carnegie Mellon University.} \quad
    Zhi-Qi Cheng\textsuperscript{2}\thanks{Corresponding Author. Now Asst. Prof., University of Washington.} \quad
    Gabriel Moreira\textsuperscript{2} \quad
    Jiawen Zhu\textsuperscript{3} \\
    Jingdong Sun\textsuperscript{2} \quad
    Bukun Ren\textsuperscript{2} \quad
    Jun-Yan He\textsuperscript{4} \quad
    Qi Dai\textsuperscript{5} \quad
    Xian-Sheng Hua\textsuperscript{1}  \vspace{0.2em} \\
    \textsuperscript{1}Zhejiang University \quad
    \textsuperscript{2}Carnegie Mellon University \\
    \textsuperscript{3}Dalian University of Technology \quad
    \textsuperscript{4}DAMO Academy, Alibaba Group \quad
    \textsuperscript{5}Microsoft Research  \vspace{0.2em} \\
\texttt{\small \{zhiqic, gmoreira, jingdons\}@cs.cmu.edu, Jiawen@mail.dlut.edu.cn, gid@microsoft.com,} \\
\texttt{\small \{jianghaoyu0608, bukunren46, junyanhe1989, huaxiansheng\}@gmail.com} \\
}
\maketitle
%%%%%%%%% ABSTRACT
\begin{abstract}
Universal Cross-Domain Retrieval (UCDR) retrieves relevant images from unseen domains and classes without semantic labels, ensuring robust generalization. Existing methods commonly employ prompt tuning with pre-trained vision-language models but are inherently limited by static prompts, reducing adaptability. We propose \emph{UCDR-Adapter}, which enhances pre-trained models with adapters and dynamic prompt generation through a two-phase training strategy. First, \emph{Source Adapter Learning} integrates class semantics with domain-specific visual knowledge using a \emph{Learnable Textual Semantic Template} and optimizes \emph{Class and Domain Prompts} via momentum updates and dual loss functions for robust alignment. Second, \emph{Target Prompt Generation} creates dynamic prompts by attending to masked source prompts, enabling seamless adaptation to unseen domains and classes. Unlike prior approaches, UCDR-Adapter dynamically adapts to evolving data distributions, enhancing both flexibility and generalization. During inference, only the image branch and generated prompts are used, eliminating reliance on textual inputs for highly efficient retrieval.
Extensive benchmark experiments show that UCDR-Adapter consistently outperforms ProS in most cases and other state-of-the-art methods on UCDR, U$^{\mathrm{c}}$CDR, and U$^{\mathrm{d}}$CDR settings.\footnote{Project:~\url{https://github.com/fine68/UCDR2024}.}
\end{abstract}

%%%%%%%%% BODY TEXT
\begin{figure}[t]
  \centering
   \includegraphics[width=0.95\linewidth]{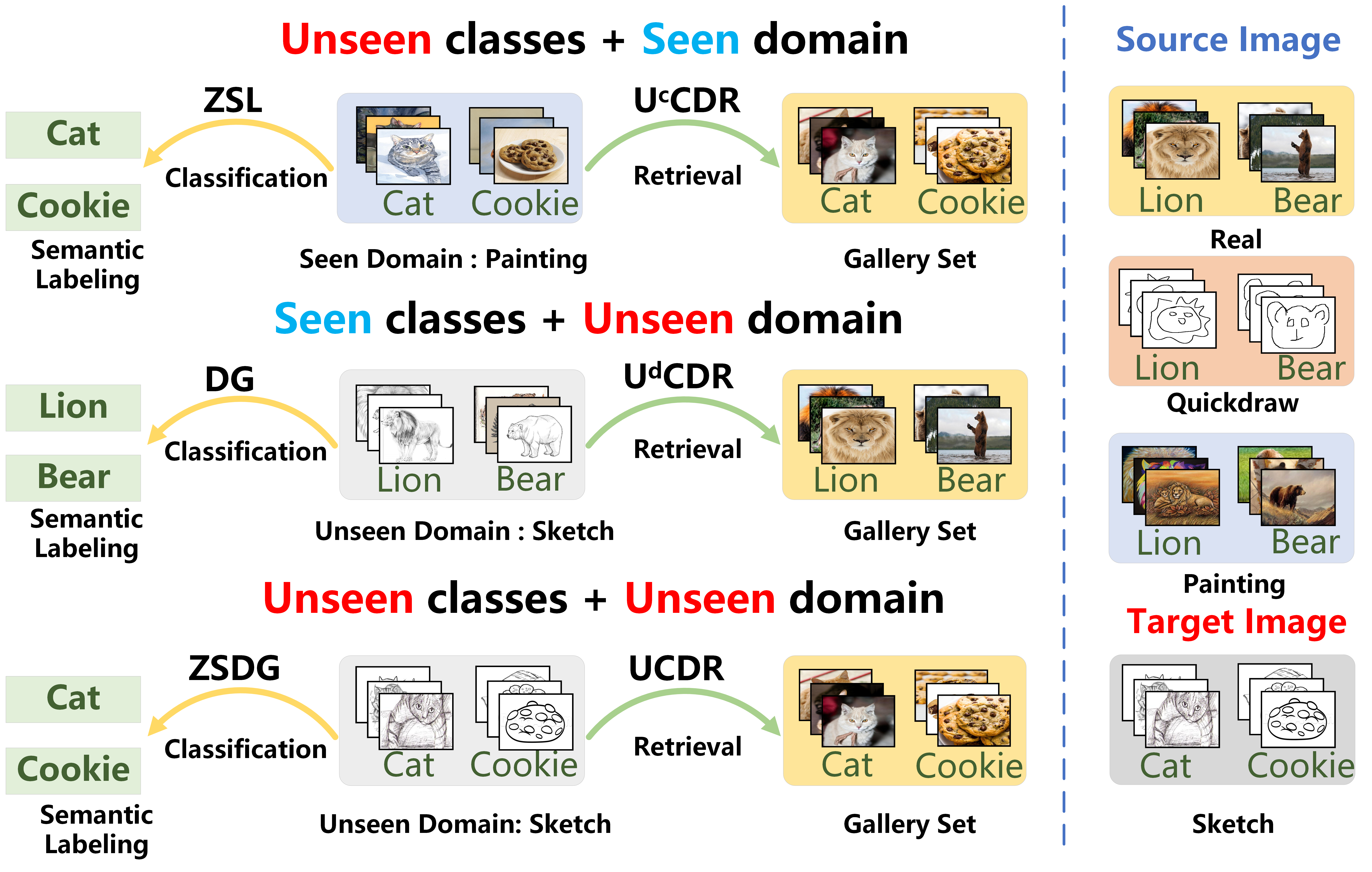}
   \vspace{-2mm}
   \caption{\small Overview of UCDR settings. Training involves seen categories (e.g., Lion) and domains (e.g., Real). Testing includes unseen domains (e.g., Sketch) and categories (e.g., Cookie) using U$^d$CDR and U$^c$CDR principles. Unlike Zero-Shot Domain Generalization (ZSDG)~\cite{mondal2022seic,mangla2022cocoa,arfeen2022handling}, UCDR does not rely on true labels for unseen data, aligning better with real-world scenarios.}
   \vspace{-0.2in}
   \label{fig:ucdr_overview}
\end{figure}

\vspace{-0.2in}
\section{Introduction}
\label{sec:intro}
Universal Cross-Domain Retrieval (UCDR)~\cite{paul2021universal} retrieves relevant images from unseen domains and classes without relying on semantic labels, ensuring robust generalization across diverse and dynamic scenarios. It addresses the challenge of mismatched training and testing data distributions~\cite{tian2022structure}, common in real-world deployments where new domains emerge during inference.~Achieving effective UCDR requires learning image representations that are both domain-agnostic and semantically discriminative~\cite{mondal2022seic}. Balancing these objectives is critical for enabling models to generalize to unseen domains and categories.

Recent approaches have leveraged pre-trained vision-language models such as CLIP~\cite{radford2021learning} and BLIP~\cite{li2022blip} to infuse semantic priors that enhance generalization~\cite{sain2023clip, pei2023clipping}. The rich world knowledge in these models helps distinguish fine-grained categories critical for retrieval. However, fine-tuning on the target dataset is often ineffective, as performance remains limited by the scarcity and lack of diversity in training data. Such data lacks sufficient coverage of the complex visual world, leading to poor generalization. Effective strategies are needed to fully leverage the knowledge in pre-trained models without being restricted by target training data limitations. This motivates advanced approaches to adapt pre-trained model priors for the UCDR task while overcoming training data constraints. 

To address these limitations, we propose \emph{UCDR-Adapter}, a comprehensive framework that enhances pre-trained models with adapter modules and dynamic prompt generation through a novel two-phase training strategy. In the first phase, \textit{Source Adapter Learning}, we integrate class semantics with domain-specific visual knowledge using a \textit{Learnable Textual Semantic Template}. We further optimize \textit{Class and Domain Prompts} via momentum-based updates and dual loss functions to achieve robust and consistent multimodal alignment. In the second phase, \textit{Target Prompt Generation}, we dynamically generate adapted prompts by attending over masked source prompts, effectively simulating adaptation to unseen domains and classes.

Unlike existing methods such as ProS~\cite{fang2024pros}, which rely on static prompts, UCDR-Adapter dynamically adjusts to evolving data distributions, improving both flexibility and generalization. By leveraging dynamic momentum-based updates, our approach effectively captures and adapts to diverse retrieval scenarios, avoiding the constraints of fixed prompt configurations. During inference, only the image branch and generated target prompts are used, eliminating reliance on textual inputs and ensuring efficient retrieval. This design enables the model to fully utilize the rich knowledge encoded in pre-trained models while overcoming the limitations of static prompts and data scarcity.

\noindent Our main contributions are summarized as follows:
\vspace{-0.1in}
\begin{enumerate}[leftmargin=*]
    \item We propose \emph{UCDR-Adapter}, a framework that enhances pre-trained vision-language models with \emph{adapter modules} and \emph{dynamic prompt generation}, adapting model knowledge for universal cross-domain retrieval tasks.
    \vspace{-0.1in}
    \item We introduce a \emph{two-phase training strategy} that integrates \emph{class semantics} with \emph{domain-specific visual knowledge} and enables dynamic prompt generation for effective adaptation to unseen domains and classes.
    \vspace{-0.1in}
    \item To achieve \emph{robust multimodal alignment}, we employ \emph{momentum-based updates} and \emph{dual loss functions}, addressing challenges such as data scarcity and limited adaptability to diverse data distributions.
    \vspace{-0.1in}
   \item Extensive experiments on benchmark datasets show that UCDR-Adapter achieves superior performance over state-of-the-art methods in most scenarios, while maintaining efficiency and offering a more robust, adaptable solution for diverse retrieval challenges.
\end{enumerate}

\begin{figure*}[t]
  \centering
   \includegraphics[width=0.9\linewidth]{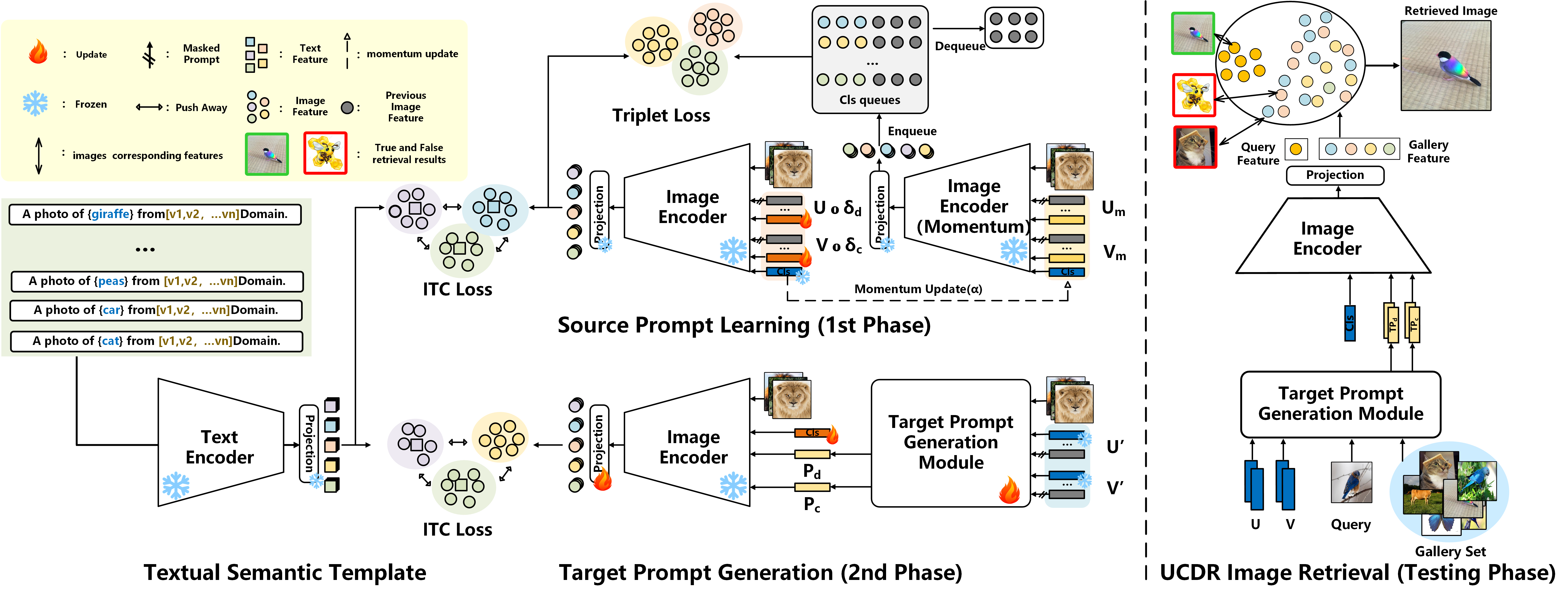}
   \vspace{-2mm}
   \caption{\small UCDR-Adapter architecture. In Phase 1 (top), \textit{Source Adapter Learning} optimizes class and domain prompts via a momentum encoder and dual loss functions for aligned multimodal representations. Only relevant prompts are activated based on the input image. In Phase 2 (bottom), the \textit{Target Prompt Generation module} generates adapted prompts by attending over masked source prompts, simulating adaptation to unseen domains and classes. At test time (right), only the image branch is utilized with the generated target prompts for effective retrieval without textual cues.} 
   \label{fig:model}
   \vspace{-4mm}
\end{figure*}

\vspace{-0.2in}
\section{Related Work}
\label{sec:related_work}
\noindent \textbf{UCDR Overview:}~Universal Cross-Domain Retrieval (UCDR)~\cite{paul2021universal} retrieves relevant images from unseen domains and classes without semantic labels, addressing mismatched training and testing distributions~\cite{tian2022structure}. This challenge is common in real-world deployments where new domains emerge during inference~\cite{cheng2016video,cheng2017video,cheng2017video2shop,cheng2017selection,nguyen2017vireo}. Effective UCDR requires learning image representations that are both domain-agnostic and semantically discriminative~\cite{mondal2022seic}, ensuring robust generalization to unseen scenarios.

\begin{table}[!t]
\small
\centering
\vspace{-2mm}
\caption{\small Comparison of ZSL, DG, U$^\mathrm{c}$CDR, U$^\mathrm{d}$CDR, ZSDG, and UCDR task settings for cross-domain retrieval.}
\vspace{-2mm}
\label{tab:comparison}
\resizebox{0.8\linewidth}{!}{%
\begin{tabular}{@{}lccl@{}}
\toprule
Task  & Query (Domain) & Query (Class) & Objective       \\ 
\midrule
ZSL \cite{blanchard2011generalizing,zhou2022domain,wang2022generalizing,xu2021fourier}   & \checkmark & \ding{55} & Classify        \\
DG \cite{gokhale2023improving,ren2023visual,cao2023review,xie2022towards}    & \ding{55} & \checkmark & Classify        \\
ZSDG \cite{mondal2022seic,mangla2022cocoa,arfeen2022handling}  & \ding{55} & \ding{55} & Classify        \\
U$^\mathrm{c}$CDR \cite{paul2021universal,tian2022structure,agarwal2023contrastive} & \checkmark & \ding{55} & Retrieve         \\
U$^\mathrm{d}$CDR \cite{paul2021universal,tian2022structure,agarwal2023contrastive} & \ding{55} & \checkmark & Retrieve         \\
UCDR \cite{paul2021universal,tian2022structure,agarwal2023contrastive}  & \ding{55} & \ding{55} & Retrieve         \\
\bottomrule
\end{tabular}%
}
\vspace{-2mm}
\end{table}

\noindent \textbf{Existing UCDR Methods:}~Various methods address UCDR. SnMpNet~\cite{paul2021universal} uses Mixup~\cite{zhang2018mixup} to blend training data at image and feature levels to simulate unseen domains and classes but requires extensive hyperparameter tuning~\cite{guo2019mixup,yu2021mixup}. SCNNet~\cite{agarwal2023contrastive} combines local feature vectors to characterize unseen data~\cite{kim2018interpretability}, yet both SnMpNet and SCNNet enforce similar representations for dissimilar domains, limiting generalization. Additionally, these methods neglect global structural information in cross-domain alignment, potentially compromising performance. SASA~\cite{tian2022structure} leverages semantic knowledge from a pre-trained visual transformer to guide a student network's fine-tuning but risks overfitting to training data, hindering generalization. Recently, ProS~\cite{fang2024pros} extended prompt tuning to UCDR by introducing  Content-aware Dynamic Prompts adapting pre-trained vision-language models like CLIP to new domains and classes.
However, the reliance on fixed-distribution data generation prompts constrains its adaptability to varying data distributions, thereby diminishing its performance across diverse scenarios.

\noindent \textbf{Vision and Language Pretraining:}~Contrastive Vision-Language Pre-training (VLP)~\cite{wang2023image,yu2022coca,li2021align} enhances visual understanding by linking text and image modalities, enabling robust semantic representations and zero-shot generalization. Early works~\cite{li2020oscar,chen2020uniter} utilized object detection models aligned with text encoders like BERT~\cite{kenton2019bert}. The CLIP architecture~\cite{radford2021learning} further advanced this by training separate visual and text encoders through contrastive learning, achieving strong adaptation to unseen classes.

\noindent \textbf{Adapters and Prompts in VLP:}~Adapter-based and prompt-tuning methods effectively adapt pre-trained vision-language models to new tasks~\cite{zhou2022learning,xie2023ra,ge2023domain,oh2024robust,oh2024towards}. Prompt tuning~\cite{petroni2019language} introduces task-specific parameters in the input while keeping the backbone frozen, enhancing downstream performance~\cite{lester2021power,li2021prefix}. Zhou \textit{et al.}~\cite{zhou2022learning} developed learnable continuous vectors to model contextual words and obtain task-related adapters. Visual Prompt Tuning (VPT)~\cite{jia2022visual} extended this to vision tasks by introducing visual adapters within the ViT architecture. Recent works like STYLIP~\cite{bose2024stylip} and CoCoOp~\cite{zhou2022conditional}, along with others~\cite{khattak2023maple,khattak2023self}, achieve domain generalization through style- and content-conditional prompt learning. However, these methods lack class-specific specialization, leading to semantic ambiguities and inadequate handling of domain gaps.

\section{Problem Definition}
\label{sec:problem_statement} 
The Universal Class-Domain Retrieval (UCDR) task aims to retrieve images from new domains and categories not seen during training. It consists of two key phases:

\vspace{1mm}
\noindent \textbf{Training on \emph{Seen} Domains~\&~Classes:}~In the training phase, we have a set of domains $\mathcal{D}_{tr}$ with $|\mathcal{D}_{tr}|$ domains. Within each domain $d \in \mathcal{D}_{tr}$, there are $N_d$ image samples. We also have a discrete set of classes $\mathcal{C}_{tr}$ with $|\mathcal{C}_{tr}|$ categories. Typically, the training data $\mathcal{S}_{tr}$ contains images across these domains and classes. The training data $\mathcal{S}_{tr}$ is {\footnotesize $\{(\mathrm{IMG}_i^{c,d}, \mathrm{TEP}_i^{c,d}, \mathrm{CLS}_i^{c,d})\}_{i=1}^{N_d}$}, where $i$ ranges from 1 to $N_d$, domain $d \in \mathcal{D}_{tr}$ and class $c \in \mathcal{C}_{tr} $. Here, {\footnotesize $\mathrm{IMG}_i^{c,d}$} is $i^{th}$ image in domain $d$ of class $c$, {\footnotesize $\mathrm{TEP}_i^{c,d}$} is the corresponding text template, and {\footnotesize $\mathrm{CLS}_i^{c,d}$} is the category label, respectively.

\vspace{1mm}
\noindent \textbf{Testing on \emph{Unseen} Domains~\&~Classes:}~At test time, the model sees new queries from a test domain set $\mathcal{D}_{te}$ not encountered during training. The test set $\mathcal{S}_{te}$ contains $N_t$ tuples {\footnotesize $\{(\mathrm{IMG}_i^{c,d}, \mathrm{CLS}_i^{c,d})\}_{i=1}^{N_t}$}, where $c \in \mathcal{C}_{te}$ and $d \in \mathcal{D}_{te}$. The key challenge is to retrieve images from a gallery matching the query image categories. 

Particularly, the UCDR task has two scenarios:
\begin{enumerate}[leftmargin=0.75cm]
\vspace{1mm}
\item \textbf{U$^{\mathrm{d}}$CDR}: The test domain $d \in \mathcal{D}_{te}$ is novel ($\mathcal{D}_{te} \not\in \mathcal{D}_{tr}$), but the classes are seen in training ($\mathcal{C}_{tr} = \mathcal{C}_{te}$).
\vspace{1mm}
\item \textbf{U$^{\mathrm{c}}$CDR}: The test domain is seen ($d \in \mathcal{D}_{tr}$), but the classes $\mathcal{C}_{te}$ are new ($\mathcal{C}_{te} \cap \mathcal{C}_{tr} = \varnothing$). 
\vspace{1mm}
\end{enumerate}
The key challenge is when both domains and classes are new, testing the limits of retrieval generalization.

\section{Proposed UCDR-Adapter}
The UCDR-Adapter enables universal cross-domain image retrieval through a three-phase approach utilizing prompt optimization: 1)~In the Source Prompt Learning phase (Sec.~\ref{sec:source_adapter_learning}), class- and domain-specific prompts are optimized via dual losses to align multimodal representations. This equips the model with source knowledge.~2)~The Target Prompt Generation phase (Sec. \ref{sec:target_prompt_generation}) produces adapted prompts by attending over masked source prompts. This simulates generalization to new domains and classes.~3) At test time (Sec. \ref{sec:ucdr_retrieval}), only the image branch is used with generated target prompts for retrieval without textual cues. By unifying strategic prompt optimization and adaptive prompt generation, UCDR-Adapter provides an efficient and effective approach for generalized UCDR.
\label{sec:UCDR_adapter}
\subsection{Source Prompt Learning (1st Phase)}
\label{sec:source_adapter_learning}
The first phase is designed to optimize class- and domain-specific prompts within the image encoder. This enables focused representation learning on the seen classes and domains encountered during training. The key innovations in the first phase are listed as:

\vspace{1mm}
\noindent \textbf{Textual Semantic Template:}~We propose a learnable semantic template to enhance cross-domain generalization, as shown in Figure \ref{fig:model}. The template is inspired by Zhou \textit{et al.}~\cite{zhou2022learning} and formulated as `\textit{A photo of \underline{$c_i$} from the \underline{$v_1,\ldots,v_{N}$} domain}', consisting of:
\begin{enumerate} [leftmargin=0.75cm]
\vspace{1mm}
\item \textbf{Class Strings $c_i$:} Precisely defined strings representing each class $c_i$ (e.g. `dog' for a image from the dog class). This integrates class semantics.
\vspace{1mm}
\item \textbf{Domain Vectors $v_j$:}~$N$ trainable $d$-dimensional vectors that capture nuanced visual characteristics of each domain. They are initialized randomly and optimized during training to represent domain-specific details (e.g. textures for the real domain).
\vspace{1mm}
\end{enumerate} 
Generally, image features from the encoder are projected into the $v_j$ space and used to update the vectors via cross-modal alignment (as Eq.~\ref{eq:itc}). This guides them to capture domain visual details. Integrating $v_j$ into the template input enables the text encoder to generate domain-aware embeddings, improving generalization.~By jointly modeling class semantics and domain visual knowledge, the template equips the model with top-down and bottom-up representations. This unified dual knowledge enhances cross-domain generalization, allowing transfer to new classes.

\vspace{1mm}
\noindent \textbf{Domain~\&~Class Prompts:}~We introduce learnable domain and class-specific prompts to incorporate domain and class semantics into the image encoder:
\begin{enumerate} [leftmargin=0.75cm]
\vspace{1mm}
\item \textbf{Domain Prompts} $U \in \mathbb{R}^{D_{tr} \times m}$: $U$ is a learnable prompt matrix with $|\mathcal{D}_{tr}|$ rows corresponding to the number of training domains, and $m$ columns representing the prompt dimensionality.
\vspace{1mm}
\item \textbf{Class Prompts} $V \in \mathbb{R}^{C_{tr} \times m}$: $V$ is a learnable prompt matrix with $|\mathcal{C}_{tr}|$ rows corresponding to the number of training classes, and $m$ columns representing the prompt dimensionality.
\vspace{1mm}
\end{enumerate}
Particularly, for an image {\footnotesize $\mathrm{IMG}_i^{c,d}$} belonging to class $c$ and domain $d$, the masks $\delta_c$ and $\delta_d$ are applied to $V$ and $U$ respectively to extract the $c^{th}$ and $d^{th}$ rows.~The masked row vectors are concatenated channel-wise into a single prompt vector $p = [V \circ \delta_c; U \circ \delta_d]$. This prompt vector $p$ is then projected to the input dimensionality of the vision transformer encoder and add to the input image features.

This allows the vision transformer to adapt its representations specifically based on the domain and class via the prompt, enhancing domain and class discrimination. The prompts are optimized via backpropagation along with the vision transformer encoder parameters to learn effective domain- and class-specific representations. By learning dedicated prompts for each domain and class, the model gains specialized representation capabilities targeted for cross-domain recognition.

\vspace{1mm}
\noindent \textbf{Momentum Updated Prompts:}~To improve sample diversity and representation learning of the class-specific prompts, we incorporate momentum updated prompts inspired by MoCo~\cite{He_2020_CVPR}: the queue acts as a large dictionary, whose size is decoupled from the batch size, allowing it to be large. Given more samples, we ensure a better coverage of the data distribution, hence the sample diversity argument.~Specifically, the momentum prompts $U_m, V_m$ are separate matrices for domain and class that are updated with momentum using the current prompt weights $U, V$ as follows:
\label{sec:moment}
\begin{equation}
\theta_M(t+1) = (1-\alpha)\theta_M(t) + \alpha\theta_C(t),  
\end{equation}
where $\theta_M, \theta_C$ are parameters of momentum ($U_m, V_m$) and current prompts ($U$, $V$). $\alpha$ is the momentum update rate, set to 0.001. And $t$ represents the number of iteration rounds.
Image features which extracted using $U_m, V_m$,  are stored in class-specific queues $\mathcal{Q} = \{q_i\}_{i=1}^{|\mathcal{C}_{tr}|}$ (see Figure~\ref{fig:model}). 

For image {\footnotesize $\mathrm{IMG}_i^{c,d}$}, positive $I_p$ and negative $I_n$ pairs are sampled from positive queue $q_c$ and negative queues $q_{-c}$. 
With the image features {\small $I_f^c$}, extracted using $p$.
And $r$ is the number of ternary loss pairs matched for $\mathrm{IMG}_i^{c,d}$, and b is margin, we set it to 0.5. The triplet loss is:
\begin{equation}
\mathcal{L}_{\mathrm{Triplet}} = \frac{1}{r} \sum_{i=1}^r \max(0, \|I_f^c - I_p\|^2 - \|I_f^c - I_n\|^2 + b),
\end{equation}
where the updated prompts increase the diversity of class representations. In short, the storing momentum features in class queues allows the sampling of hard positives and negatives for robust learning. {\small $\mathcal{L}_\mathrm{Triplet}$} loss on these samples enhances inter-class discrepancy and intra-class similarity.

\vspace{1mm}
\noindent \textbf{Cross-Modal Alignment:}~Cross-modal alignment is used to align image and text representations. This process helps to improve the domain invariance of image features and optimize the prompts. We use a contrastive loss to align the encoded image features ($I_f^c$) for class $c$ with the encoded text features ($T_f^c$) from the semantic template. The {\small $\mathcal{L}_{\mathrm{ITC}}$} loss is defined as:
\begin{equation}
\mathcal{L}_{\mathrm{ITC}} = -\log \frac{\exp(s(I_f^c, T_f^c)/\tau)}{\sum_{k=1}^{|\mathcal{C}_{tr}|} \exp(s(I_f^c, T_f^k)/\tau)}, 
\label{eq:itc}
\end{equation}
where $s$ represents cosine similarity and $\tau$ is temperature, set to 0.07. The {\small $\mathcal{L}_{\mathrm{ITC}}$}  function brings the matched image and text pairs closer and pushes non-matched pairs apart.

This alignment between image and text features helps the model learn domain-agnostic representations. By aligning image and text features across domains, the model learns domain-agnostic representations.~The $\mathcal{L}_\mathrm{ITC}$ loss finetunes the prompts and semantic templates to align with cross-modal semantics.

\begin{equation}
\mathcal{L}_{\mathrm{Phase1}} = \mathcal{L}_{\mathrm{Triplet}} + \mathcal{L}_{\mathrm{ITC}}, 
\label{eq:stage1}
\end{equation}
the overall loss in Phase1({\small $\mathcal{L}_{\mathrm{Phase1}}$}) combines  {\small$\mathcal{L}_{\mathrm{ITC}}$} and  {\small $\mathcal{L}_{\mathrm{Triplet}}$}.~In summary, cross-modal alignment provides dual-supervision from both visual and semantic modalities to improve domain invariance and prompt optimization.

\begin{figure}[t]
  \centering
   \includegraphics[width=1.0\linewidth]{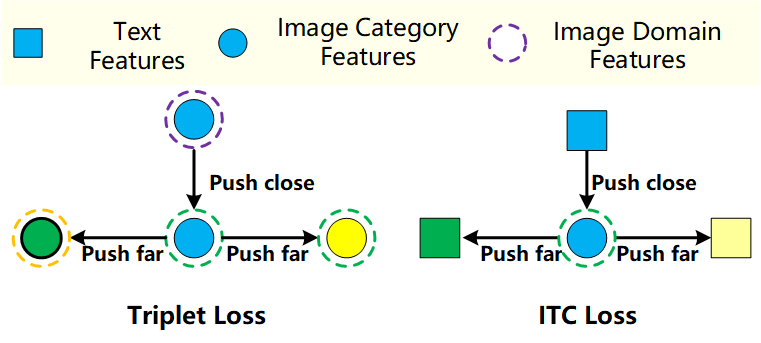}
   \caption{Phase 1 loss function. Right: ITC loss. Left: Triplet Loss.~Solid circles are image features, boxes represent the semantic features of each class, and the dashed circles correspond to the image domain.}
   \vspace{-4mm}
   \label{fig:stage1_loss}
   \vspace{-2mm}
\end{figure}

\subsection{Target Prompt Generation (2nd Phase)}
\label{sec:target_prompt_generation}
\begin{figure}[t]
  \centering
   \includegraphics[width=0.8\linewidth]{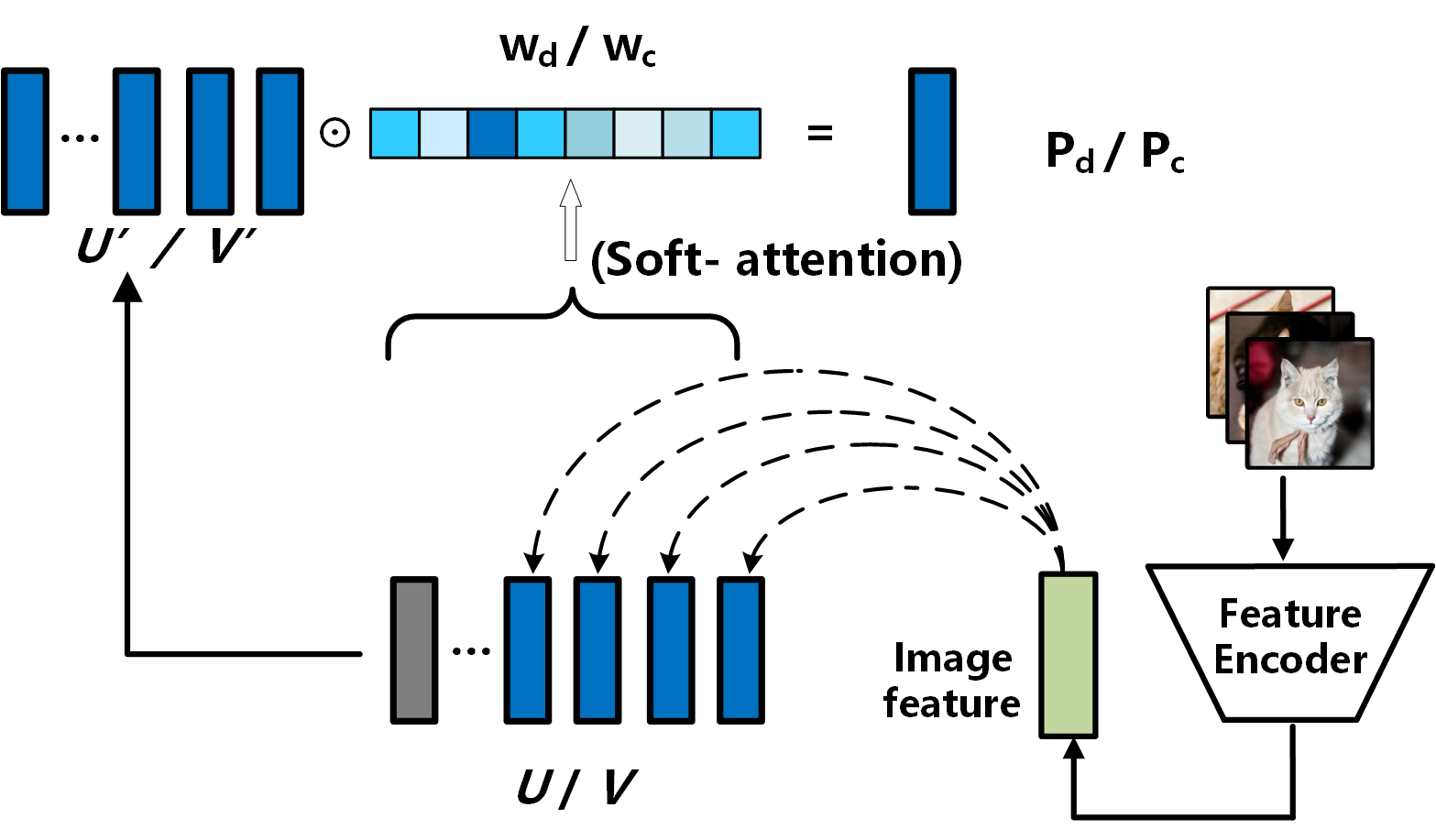}
   \vspace{-2mm}
   \caption{Target Prompt Generation process. Gray indicates that the class and domain prompts are masked. Where ${P}_d$ and ${P}_c$ are target prompts generated for unseen domains and classes.}
   \label{fig:tpg}
   \vspace{-2mm}
\end{figure}

In the second phase, we simulate a scenario where the class and domain of an image are unknown, to generate target prompts to achieve generalization to new domains and classes. As illustrated in Figure \ref{fig:tpg}, the Target Prompt Generation (TPG) module leverages the learned prompts from phase 1.
We mask out the rows of $U,V$ corresponding to the known domains and classes using $1-\delta_c$ and $1-\delta_d$:
\begin{align}
\vspace{-2mm}
U' &= U \circ (1-\delta_d), \\
V' &= V \circ (1-\delta_c), 
\vspace{-4mm}
\end{align}
this leaves us with prompts with masked-out familiar domains ($U'$) and classes ($V'$).

Particularly, the image {\small {$\mathrm{IMG}_i^{c,d}$}} is passed through the Feature Encoder $g()$ to extract features $I_g \in \mathbb{R}^{D}$. And Attn() represent Soft-Attention. We compute attention weights over the masked prompts:
\begin{align}
\vspace{-2mm}
w_d &= \text{Attn}(I_g, U'),\\
w_c &= \text{Attn}(I_g, V'), 
\vspace{-2mm}
\end{align}
this aligns the image features with the unfamiliar prompt rows. The attention weights ($w_d$ and $w_c$) are used to compute target prompts as weighted combinations of the masked rows as $P_d = V' w_d$ and 
$P_c = U' w_c$.~Here the target prompts are concatenated and projected to the encoder input dimension, then add to the input image features for enhanced feature extraction. The extracted features $I_f^c$ are aligned with $T_f^c$ from phase 1 via $\mathcal{L}_\mathrm{ITC}$ to learn generalized representations. By generating target prompts via attention over masked prompts, the model learns to produce prompts for new domains and classes.

\subsection{UCDR Image Retrieval~(Testing Phase)}
\label{sec:ucdr_retrieval}
In the testing phase for UCDR task, textual semantic information is unavailable for new classes. Thus, we discard the textual components and only utilize the image branch. The key components are:
\begin{enumerate} [leftmargin=0.75cm]
\vspace{1mm}
\item \textbf{Domain~\&~Class Prompts} from phase 1 - the domain and class prompts for seen classes/domains.
\vspace{1mm}
\item \textbf{Target Prompt Generation  Module} from phase 2 to generate adapted prompts.
\vspace{1mm}
\item \textbf{Frozen Image Encoder} from the pre-trained CLIP model does not require any extra modifications.
\vspace{1mm}
\end{enumerate}
For a query image and each gallery image, the TPG module is applied to create adapted target prompts based on the image content. These target prompts are input to the image encoder to extract enhanced representations for each image.~The query embedding is then compared to the gallery embeddings via Euclidean distance, and the gallery images are ranked based on distance to retrieve the final results. The same process is applied for the U$^c$CDR and U$^d$CDR sub-tasks without modification, utilizing the TPG capability to generalize. By retaining only the crucial image branch with the prompt generation, the model is able to perform retrieval for new classes and domains at test time without textual cues.

\section{Experiments}
\subsection{Experimental Setup}
\noindent \textbf{Datasets:}~Following the benchmarks in \cite{paul2021universal,tian2022structure,mondal2022seic}, we evaluate the performance of the UCDR-Adapter model on three widely used datasets (DomainNet \cite{peng2019moment}, Sketchy \cite{liu2017deep,sangkloy2016sketchy}, TU-Berlin \cite{liu2017deep,eitz2010evaluation}) for three retrieval tasks, namely UCDR, U$^d$CDR and U$^c$CDR. DomainNet contains 596,006 images from 345 classes and 6 domains (real, Sketch, Quickdraw, Infograph, Clipart, Painting) for U$^d$CDR and UCDR experiments. and used 245 classes as a training set, 55 classes as a validation set, and 45 classes as a test set as previously set \cite{paul2021universal}. In training, we use images from five domains, to satisfy the unseen setting. Considering that the training data for the CLIP model consists of image-text pairs downloaded from the web, we do not use the Real as the unseen domain. In addition, we have two search settings: 1)~Use an unseen class image belonging to the real domain as the gallery set;~2)~Considering more realistic scenarios, the gallery set is a combination of visible and invisible classes from the real domain. Sketchy contains 75,471 sketches and 73,002 images from 125 categories. Based on the settings in \cite{paul2021universal,tian2022structure,mondal2022seic}, the training set, validation set, and test set consist of 93, 11, and 21 categories, respectively. Sketchy and TU-Berlin are both used for U$^c$CDR. 

\begin{table*}
    \renewcommand*{\arraystretch}{1.1}
    \footnotesize
  \centering
  % \begin{tabular}{@{}lc@{}}
\caption{Evaluation results on DomainNet for UCDR using two different gallery settings: (1) Only unseen class samples from the held-out domain; (2) Both seen and unseen class samples from the held-out domain. The results demonstrate UCDR-Adapter's effectiveness in retrieving unseen classes under domain shift.}
\resizebox{0.8\linewidth}{!}{%
  \begin{tabular}{c@{}c@{}c@{}c@{}c@{}c@{}c@{}}
    \toprule
    % Training Domains& Query Domain& Method & \multicolumn{2}{c}{unseen class gallery}& \multicolumn{2}{c}{seen+unseen class gallery}\\
    \multirow{2}{*}{Training Domains} & \multirow{2}{*}{Query Domain} & \multirow{2}{*}{Method} & \multicolumn{2}{c}{Unseen Class Gallery} & \multicolumn{2}{c}{Seen+Unseen Class Gallery}\\
    \cline{4-5} \cline{6-7}
    & & & mAP@200 & Prec@200 & mAP@200 & Prec@200\\
    \midrule
    \multirow{6}{*}{\parbox{4cm}{\centering Real, Quickdraw,\\Infograph, Painting,\\Clipart}} &  \multirow{6}{*}{ Sketch} & SnMpNet [ICCV 2021] \cite{paul2021universal} & 0.3007 & 0.2432 & 0.2624 & 0.2134 \\
    & & SCNNet [WACV 2023] \cite{agarwal2023contrastive} & 0.4075 & 0.4120 & 0.3422 & 0.2534 \\
    & & SASA [SIGIR 2022] \cite{tian2022structure} & 0.5262 & 0.4468 & 0.4732 & 0.4025 \\
    & & Zero-Shot CLIP   & 0.4222 & 0.3529 & 0.3760 & 0.3081 \\
    & & ProS [CVPR 2024]\cite{fang2024pros}  & 0.6457 & 0.6001 & 0.5843 & 0.5463 \\
    & & \textbf{UCDR-Adapter(Ours)} & \textbf{0.6591} & \textbf{0.6142} & \textbf{0.6073} & \textbf{0.5707} \\
    \midrule
    \multirow{6}{*}{\parbox{4cm}{\centering Real, Sketch,\\ Infograph, Painting,\\ Clipart}} & \multirow{6}{*}{Quickdraw} & SnMpNet [ICCV 2021] \cite{paul2021universal} & 0.1736 & 0.1284 & 0.1512 & 0.1111 \\
    & & SCNNet [WACV 2023] \cite{agarwal2023contrastive} & 0.1998 & 0.1580 & 0.1698 & 0.1411 \\
    & & SASA [SIGIR 2022] \cite{tian2022structure} & 0.2564 & 0.1970 & 0.2116 & 0.1651 \\
    & & Zero-Shot CLIP   & 0.0744 & 0.0561 & 0.0607 & 0.0386 \\
    & & ProS [CVPR 2024]\cite{fang2024pros}   & \textbf{0.2842} & \textbf{0.2544} & \textbf{0.2318} & 0.2127 \\
    & & \textbf{UCDR-Adapter(Ours)} & 0.2794 & 0.2534 & 0.2317 & \textbf{0.2154} \\
    \midrule
    \multirow{6}{*}{ \parbox{4cm}{\centering Real, Sketch,\\ Quickdraw, Infograph,\\ Clipart}} & \multirow{6}{*}{Painting} & SnMpNet [ICCV 2021] \cite{paul2021universal} & 0.4031 & 0.3332 & 0.3635 & 0.3019 \\
    & & SCNNet [WACV 2023] \cite{agarwal2023contrastive} & 0.4242 & 0.4409 & 0.3731 & 0.3964 \\
    & & SASA [SIGIR 2022] \cite{tian2022structure} & 0.5898 & 0.5188 & 0.5463 & 0.4804 \\
    & & Zero-Shot CLIP   & 0.6169 & 0.5508 & 0.5755 & 0.5085 \\
    & & ProS [CVPR 2024]\cite{fang2024pros}   & 0.7516 & 0.6955 & 0.7120 & 0.6612 \\
    & & \textbf{UCDR-Adapter(Ours)} & \textbf{0.7538} & \textbf{0.6974} & \textbf{0.7203} & \textbf{0.6693} \\
    \midrule
    \multirow{6}{*}{ \parbox{4cm}{\centering Real, Sketch,\\ Quickdraw, Painting,\\ Clipart}} & \multirow{6}{*}{Infograph} & SnMpNet [ICCV 2021] \cite{paul2021universal} & 0.2079 & 0.1717 & 0.1800 & 0.1496 \\
    & & SCNNet [WACV 2023] \cite{agarwal2023contrastive} & 0.2737 & 0.2476 & 0.2369 & 0.1983 \\
    & & SASA [SIGIR 2022] \cite{tian2022structure} & 0.2823 & 0.2425 & 0.2491 & 0.2113 \\
    & & Zero-Shot CLIP   & 0.5007 & 0.4474 & 0.4501 & 0.3990 \\
     & & ProS [CVPR 2024]\cite{fang2024pros}   & \textbf{0.5798} & \textbf{0.5442} & 0.5219 & 0.4956 \\
    & & \textbf{UCDR-Adapter(Ours)} & 0.5714 & 0.5364 & \textbf{0.5315} & \textbf{0.5022} \\
    \midrule
    \multirow{6}{*}{\parbox{4cm}{\centering Real, Sketch,\\ Quickdraw, Infograph,\\ Painting}} & \multirow{6}{*}{Clipart} & SnMpNet [ICCV 2021] \cite{paul2021universal} & 0.4198 & 0.3323 & 0.3765 & 0.2959 \\
    & & SCNNet [WACV 2023] \cite{agarwal2023contrastive} & 0.4843 & 0.4664 & 0.4322 & 0.4016 \\
    & & SASA [SIGIR 2022] \cite{tian2022structure} & 0.5392 & 0.4300 & 0.4902 & 0.3886 \\
    & & Zero-Shot CLIP   & 0.6037 & 0.5131 & 0.5700 & 0.4768 \\
    & & ProS [CVPR 2024]\cite{fang2024pros}   & 0.7648 & 0.7186 & 0.7228 & 0.6815 \\
    & & \textbf{UCDR-Adapter(Ours)} & \textbf{0.7718} & \textbf{0.7263} & \textbf{0.7391} & \textbf{0.6979}\\
    \midrule

    \multicolumn{2}{c}{\multirow{6}{*}{Average}} & SnMpNet [ICCV 2021] \cite{paul2021universal} & 0.3010 & 0.2418 & 0.2667 & 0.2144 \\
    & & SCNNet [WACV 2023] \cite{agarwal2023contrastive} & 0.3579 & 0.3449 & 0.3108 & 0.2981 \\
    & & SASA [SIGIR 2022] \cite{tian2022structure} & 0.4387 & 0.3670 & 0.3940 & 0.3295 \\
    & & Zero-Shot CLIP   & 0.4435 & 0.3840 & 0.3737 & 0.3462 \\
    & & ProS [CVPR 2024]\cite{fang2024pros}   & 0.6052 & 0.5626 & 0.5546 & 0.5195 \\
    & & \textbf{UCDR-Adapter(Ours)} & \textbf{0.6071} & \textbf{0.5655} & \textbf{0.5660} & \textbf{0.5311} \\

% Continue with the rest of the table as needed

    \bottomrule
  \end{tabular}%
  }
  \label{tab:ucdr_domainnet}
  \vspace{-2mm}
\end{table*}

\noindent \textbf{Evaluation Metrics:}~To ensure a fair comparison, we use the same assessment metrics as \cite{paul2021universal,tian2022structure,mondal2022seic}. For Sketchy and DomainNet,~we evaluate the top 200 candidates' precision and mean average precision scores (Prec@200 and mAP@200).~For TU-Berlin, we used Prec@100 and mAP@all as evaluation metrics.

\noindent \textbf{Implementation Details:}~We implement UCDR-Adapter in PyTorch and train on a single Nvidia A100 GPU. ViT-B/32 is used for both image and text encoders. The Domain Prompts ($U$) and Class Prompts ($V$) are 768-d vectors.
The momentum encoder has a momentum update rate of 1e-3 and a queue length of 20 per class. Training runs for 50 epochs with early stopping if validation accuracy does not improve for 2 epochs. Stage 1 uses a batch size of 400, and stage 2 uses 50.
Adam optimization is used with initial LR 1e-3 decayed to 1e-6 over 20 epochs. For Zero-Shot CLIP, we use pre-trained VIT-B/32 Image Encoder for retrieval.

\begin{table}
\renewcommand*{\arraystretch}{1.1}
\footnotesize
\centering
\caption{U$^d$CDR evaluation results on DomainNet, where queries from holdout domain during training.}
\label{tab:udcdr}
\resizebox{0.9\linewidth}{!}{%
\begin{tabular}{c@{}c@{}c@{}c@{}}
\toprule
{Query Domain} & {Methods} & {mAP@200} & {Prec@200} \\ \midrule
\multirow{4}{*}{\parbox{2cm}{\centering Sketch }} & SnMpNet \cite{paul2021universal}  & 0.3529 & 0.1657 \\
             & SASA \cite{tian2022structure}  & 0.5733 & 0.5290 \\
             & ProS\cite{fang2024pros}   & \textbf{0.7385} & \textbf{0.4911} \\
             & \textbf{UCDR-Adapter(Ours)}  & 0.7332 & 0.4893 \\
\midrule
\multirow{4}{*}{\parbox{2cm}{\centering Quickdraw }} & SnMpNet \cite{paul2021universal}  & 0.1077 & 0.0509 \\
             & SASA \cite{tian2022structure} & 0.1805 & 0.1549 \\
             & ProS\cite{fang2024pros}   & 0.2889 & \textbf{0.1186} \\
             & \textbf{UCDR-Adapter(Ours)}  & \textbf{0.2900} & 0.1181 \\
\midrule
\multirow{4}{*}{\parbox{2cm}{\centering Painting }}    & SnMpNet \cite{paul2021universal} & 0.4808 & 0.4424 \\
             & SASA \cite{tian2022structure} & 0.5596 & 0.5178 \\
             & ProS\cite{fang2024pros}   & 0.7227 & 0.4615 \\
             & \textbf{UCDR-Adapter(Ours)}  & \textbf{0.7306} & \textbf{0.4634} \\
\midrule
\multirow{4}{*}{\parbox{2cm}{\centering Infograph }}   & SnMpNet \cite{paul2021universal} & 0.1957 & 0.1764 \\
             & SASA \cite{tian2022structure} & 0.2340 & 0.2093 \\
             & ProS\cite{fang2024pros}   & 0.6056 & \textbf{0.3962} \\
             & \textbf{UCDR-Adapter(Ours)}   & 0.\textbf{6064} & 0.3922 \\
\midrule
\multirow{4}{*}{\parbox{2cm}{\centering Clipart }}  & SnMpNet \cite{paul2021universal} & 0.5520 & 0.5074 \\
             & SASA \cite{tian2022structure} & 0.6840 & 0.6361 \\
             & ProS\cite{fang2024pros}   & 0.8105 & 0.5298 \\
             & \textbf{UCDR-Adapter(Ours)}  & \textbf{0.8251} & \textbf{0.5392} \\
\midrule
\multirow{4}{*}{\parbox{2cm}{\centering Average }}  & SnMpNet \cite{paul2021universal} & 0.3378 & 0.2685 \\
             & SASA \cite{tian2022structure} & 0.4462 & 0.4094 \\
             & ProS \cite{fang2024pros} & 0.6332 & 0.3994 \\
              & \textbf{UCDR-Adapter(Ours)}  & \textbf{0.6370} & \textbf{0.4004} \\
\bottomrule
\end{tabular}%
}
\vspace{-2mm}
\end{table}

\subsection{Comparison with State-of-the-art}
\noindent \textbf{UCDR Results:}~We evaluate UCDR-Adapter on UCDR using DomainNet. Following the protocol in~\cite{paul2021universal}, we select unseen samples from five domains for training, and use one holdout domain for querying unseen classes. We consider two galleries:~1) unseen classes only;~2) both seen and unseen classes. The results are in Table~\ref{tab:ucdr_domainnet}. UCDR-Adapter improves the mAP@200 over SASA~\cite{tian2022structure} by +16.84\% and +17.2\% on the two galleries. It also outperforms Zero-Shot-CLIP, which is comparable to SASA but degrades on less detailed domains. Compared to Zero-Shot-CLIP, UCDR-Adapter achieves gains of +16.36\% and +19.23\% respectively. This shows our training strategy enhances generalization over the pre-trained features for unseen classes and domains. The consistent improvements show UCDR-Adapter's effectiveness on the UCDR task by adapting the pre-trained knowledge through prompt optimization.

\noindent \textbf{U$^d$CDR Results:}~We evaluate UCDR-Adapter's domain generalization ability on U$^d$CDR using DomainNet.~As shown in Table~\ref{tab:udcdr}, we select 25\% queries per seen class from the unseen domain (10\% for Quickdraw given its large size).~The gallery contains seen class images from the Real domain. UCDR-Adapter improves mAP@200 over SASA by +19.08\%, indicating our domain-level adapter fine-tuning allows CLIP to focus on class-discriminative features, mitigating the impact of domain shift.~The boost shows UCDR-Adapter's ability to learn domain-invariant representations, crucial for generalization in U$^d$CDR.

\noindent \textbf{U$^c$CDR Results:}~We evaluate UCDR-Adapter's generalization to unseen classes on U$^c$CDR using Sketchy and TU-Berlin. As shown in~Table~\ref{tab:uccdr}, queries are unseen classes from the seen training domain, while the gallery contains all classes.
On Sketchy, UCDR-Adapter improves mAP@200 and prec@200 over SASA by +3.75\% and +7.98\% respectively. On TU-Berlin, it achieves gains of +18.66\% in mAP@200 and +6.35\% in Allprec@100. This demonstrates the TPG module can effectively leverage acquired knowledge to infer representations of unseen classes and domains.~These consistent improvements verify UCDR-Adapter's ability to generalize to new categories, which is crucial for the U$^c$CDR task.

\begin{table}
\renewcommand*{\arraystretch}{1.1}
\footnotesize
\centering
\vspace{-1mm}
\caption{U\textsuperscript{c}CDR evaluation results on Sketchy and TU-Berlin.}
\label{tab:uccdr}
\resizebox{0.9\linewidth}{!}{%
\begin{tabular}{c@{}c@{}c@{}c@{}c@{}}
\toprule
\multirow{2}{*}{Method} & \multicolumn{2}{c}{Sketchy} & \multicolumn{2}{c}{TU-Berlin} \\
\cmidrule(r){2-3} \cmidrule(l){4-5}
       & {mAP@200} & {prec@200} & {mAP@All} & {prec@100} \\
\midrule
SnMpNet \cite{paul2021universal}      & 0.5781 & 0.5155 & 0.3568 & 0.5226 \\
SASA \cite{tian2022structure}         & 0.6910 & 0.6090 & 0.4715 & 0.6682 \\
ProS \cite{fang2024pros}         & 0.6991 & 0.6545 & \textbf{0.6675} & \textbf{0.7442} \\
\textbf{UCDR-Adapter(Ours)}        & \textbf{0.7285} & \textbf{0.6888} & 0.6581 & 0.7317 \\
\bottomrule
\end{tabular}%
}
\vspace{-1mm}
\end{table}

\begin{figure*}[t]
  \centering
   \includegraphics[width=0.9\linewidth]{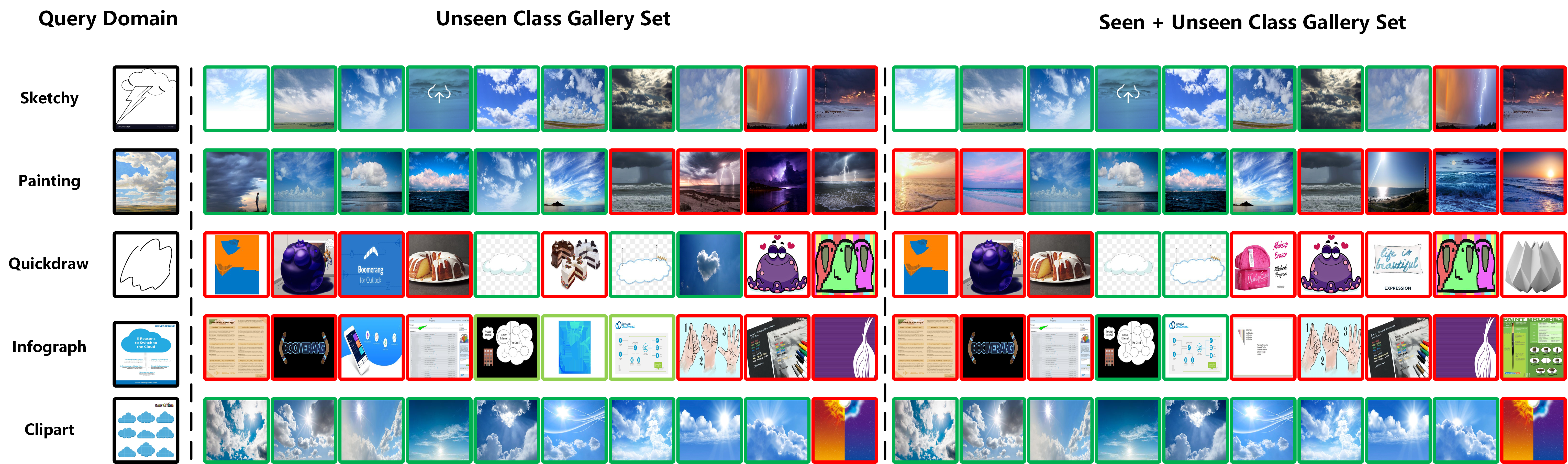}
   \caption{The results of using UCDR-Adapter for the UCDR  on DomainNet. The `Cloud' class from the holdout domain as the query.}
   \label{fig:twocolumn}
   \vspace{-3mm}
\end{figure*}

\subsection{Ablation Studies}
\label{sec:able}

We conduct ablation experiments on Sketchy as the unseen domain, with unseen classes for both queries and gallery. The results are in Table \ref{tab:transfer_methods}. A comparison of training parameters is shown in Table \ref{tab:train}.

\noindent \textbf{Fine-tuning vs Adapters:}~Fine-tuning the entire CLIP model~\cite{radford2021learning} is computationally expensive as it requires updating the whole pre-trained network. More critically, fine-tuning often leads to catastrophic forgetting, degrading the original capabilities of CLIP. An alternative is a linear probe, where only a classifier layer is trained on top of frozen CLIP features. As shown in Table~\ref{tab:transfer_methods}, the linear probe improves over zero-shot CLIP but underperforms our proposed adapter approach, which achieves a significant gain of +5.69\% in mAP. This verifies that adapters can efficiently tune CLIP for downstream tasks through the targeted updates of small adapter modules, without interfering with the frozen pre-trained weights. Adapters provide an effective balance between transferability and adaptation. Our designed prompts and training strategy take advantage of this to unlock CLIP's knowledge for UCDR.

\noindent \textbf{Two Phases vs One Phase:} Comparing adapters trained end-to-end versus generated by the TPG module, TPG improves mAP by +6.82\%. This validates our two-stage approach, where TPG harnesses knowledge from seen classes effectively through masking.

\begin{figure}[t]
  \centering
   \includegraphics[width=1.0\linewidth]{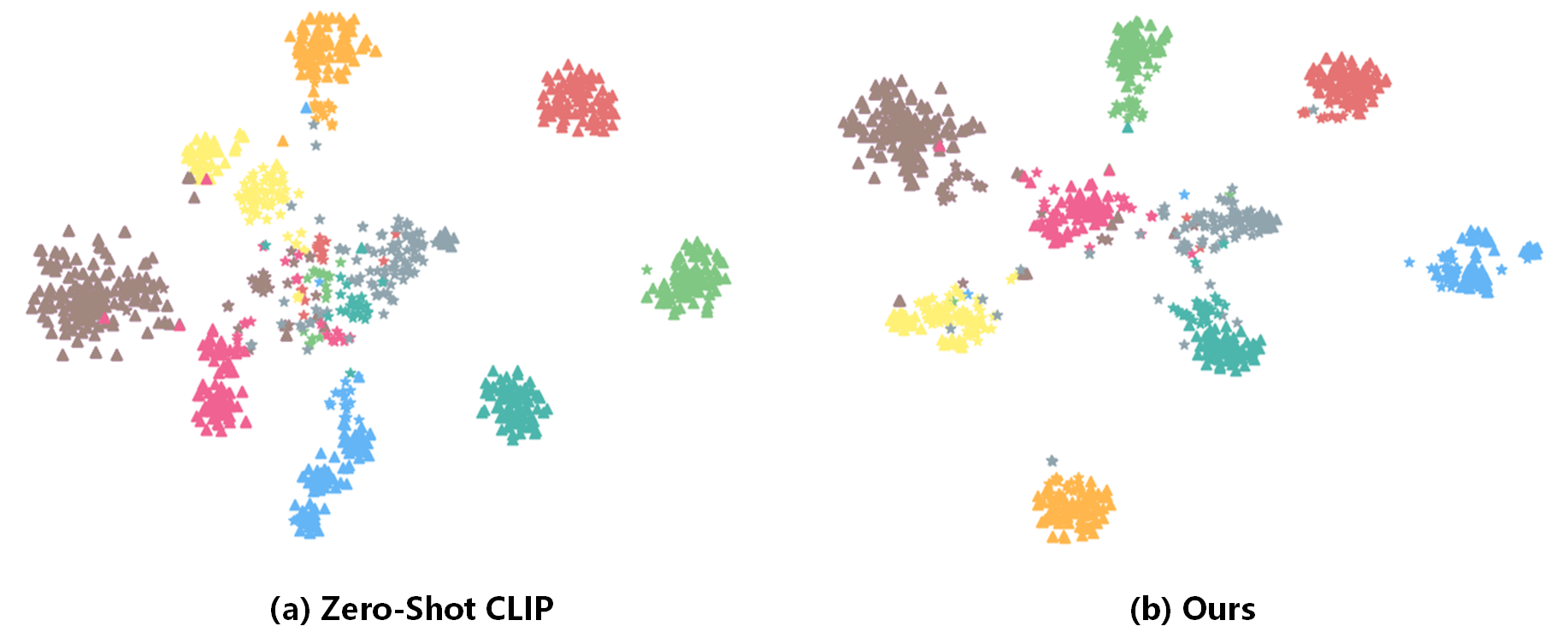}
   \vspace{-2mm}
   \caption{t-SNE visualization of features for 10 random unseen classes from Real (stars) and Sketchy (triangles) domains. Sketchy was used as the unseen holdout domain during training. Colors indicate different classes.~(a) CLIP features show entanglement between some classes.~(b) Our adapted model shows improved inter-class separation and intra-class clustering, indicating enhanced discrimination of unseen classes.}
   \label{fig:onecol}
   \vspace{-4mm}
\end{figure}

\noindent \textbf{Learnable Semantic Template:} Replacing the fixed CLIP text encoding with our learnable semantic template improves mAP by +2.87\%, confirming the benefits of optimizing domain information into category supervision.

\noindent \textbf{Momentum \& Triplet Loss:} The momentum encoder and triplet loss help enhance sample diversity and representation learning of class/domain knowledge in the adapters. The momentum encoder maintains a diverse set of class-specific samples in queues for mining hard positives/negatives. Combined with the triplet loss, this strengthens inter-class discrepancy and intra-class similarity in adapter representations. When integrated in phase 2 training, the improved adapters lead to better generalization.~Our full UCDR-Adapter achieves significant gains over zero-shot CLIP while only adding minimal parameters in the adapters.

\begin{table}
\renewcommand*{\arraystretch}{1.1}
\footnotesize
\centering
\caption{UCDR ablation experiments, with Sketchy as the holdout domain. $\mathrm{Prompt_{vision}}$ and  $\mathrm{Prompt_{text}}$ are image and text prompts, respectively. $U$ and $V$ are domain and category prompts. Mask uses two-phases masking operation, and TST is a training Textual Semantic Template. $\mathcal{L}_{\text {Triple }}$  (pair = i) is the use of $i$ ternary loss pairs.}
\label{tab:transfer_methods}
\resizebox{1.0\linewidth}{!}{%
\begin{tabular}{c@{}c@{}c@{}c@{}}
\toprule
Tuning Phases & \multirow{1}{*}{Training Task}&  mAP@200 &  Prec@200 \\
\midrule
\multirow{5}{*}{One Phase} & Fine-Tuning CLIP    & 0.2674 & 0.2063\\
& Zero-Shot CLIP & 0.4222 & 0.3529\\
& Linear Probing(LP) CLIP & 0.4936 & 0.4365\\
& LP + $\mathrm{Prompt_{vision}}$ & 0.5505 & 0.4917\\
& LP + $\mathrm{Prompt_{vision}}$ + $\mathrm{Prompt_{text}}$      & 0.5763 & 0.5202\\
\midrule
\multirow{5}{*}{Two Phases}&    $U$ + $V$ + TPG           & 0.6187 & 0.5753\\
&   $U$ + $V$ + TPG + Mask        & 0.6253 & 0.5807\\
&  Full ($U$ + $V$ + TPG + Mask + TST)           & 0.6540 & 0.6062\\
&  Full + $\mathcal{L}_{\text {Triple }}$ (pair 1)        & 0.6560 & 0.6107\\
& \textbf{Full UCDR-Adapter (Ours)}      & \textbf{0.6591} & \textbf{0.6142} \\
\bottomrule
\end{tabular}%
}
\vspace{-2mm}
\end{table}

\begin{table}
\footnotesize
\centering
\caption{Comparison of the number of parameters and mean average precision for the three tuning CLIP methods.}
\label{tab:train}
\resizebox{0.9\linewidth}{!}{%
\begin{tabular}{c@{}c@{}c@{}}
\toprule
\parbox{1.5cm}{\centering Methods} &  \parbox{2cm}{\centering Training Parameter(M)} & \parbox{2cm}{\centering mAP@200}\\
\midrule
Fine-Tuning (Image encoder)  &  87.84 &  0.2674\\
Linear Probing  &  0.39 & 0.4936\\
\textbf{UCDR-Adapter(Ours)}&  \textbf{2.36} & \textbf{0.6591}\\
\bottomrule
\end{tabular}%
}
\vspace{-5mm}
\end{table}

\noindent \textbf{Retrieval Qualitative Analysis:}~Figure \ref{fig:twocolumn} shows the top 10 retrieval results for a `cloud' query from the Sketchy, Painting, Clipart, and Quickdraw domains under the U$^{c}$CDR set. For domains with more distinctive features like Sketchy, Painting, and Clipart, the model retrieves fewer incorrect candidates compared to Quickdraw which lacks detail. In some cases, errors occur due to background similarities, as with the incorrect lightning' retrieval.

We also analyze the feature distributions of CLIP and our adapted model in Figure \ref{fig:onecol}. Using 10 random unseen classes from Real (seen) and Sketchy (unseen) domains, CLIP fails to differentiate some classes in Figure \ref{fig:onecol}(a). In contrast, our adapted CLIP in Figure \ref{fig:onecol}(b) shows improved intra-class clustering and inter-class separation. This indicates our approach enhances CLIP's discriminative power while retaining its representation capabilities, crucial for the differentiation of unseen classes.

\section{Conclusion}
We introduce \emph{UCDR-Adapter} for Universal Cross-Domain Retrieval (UCDR), enhancing pre-trained models through adapter-based prompt optimization and a two-phase training strategy. First, class- and domain-specific prompts are optimized with momentum encoders and dual loss functions for aligned multimodal representations. Then, target prompts are generated by attending to masked source prompts, facilitating adaptation to unseen domains and classes. Experiments demonstrate that UCDR-Adapter outperforms state-of-the-art methods on UCDR and its subtasks, ensuring effective generalization. This framework advances prompt-based tuning for UCDR and extends to other cross-domain tasks, offering an efficient solution for generalized visual retrieval.

\clearpage

\section*{Acknowledgments}
This work was partially supported by the Air Force Research Laboratory under agreement number FA8750-19-2-0200, and by grants from the Defense Advanced Research Projects Agency (DARPA) under the GAILA program (award HR00111990063) and the AIDA program (FA8750-18-20018). Zhi-Qi Cheng also acknowledges support from the University of Washington startup fund, the Intel Ph.D. Fellowship, and the IBM Outstanding Student Scholarship. Portions of this research were funded by these grants. The U.S. Government is authorized to reproduce and distribute reprints for governmental purposes notwithstanding any copyright notation thereon. The views and conclusions expressed in this work are those of the authors and do not necessarily reflect the official policies or endorsements of the Air Force Research Laboratory, DARPA, or the U.S. Government.

%%%%%%%%% REFERENCES
{\small
\bibliographystyle{ieee_fullname}
\bibliography{egbib}
}

\end{document}